\documentclass[letterpaper, 10 pt, conference]{ieeeconf} % Comment this line out
\IEEEoverridecommandlockouts                              % This command is only
\usepackage{epsfig} % for postscript graphics files
\usepackage{url}

\usepackage{amsmath} % assumes amsmath package installed
\usepackage{amssymb}  % assumes amsmath package installed
\usepackage{multirow}
\usepackage{verbatim}

\let\labelindent\relax
\usepackage{enumitem}% http://ctan.org/pkg/enumerate

\usepackage{multicol}        % used for the two-column index
\usepackage[bottom]{footmisc}% places footnotes at page bottom
\usepackage{algorithm}
\usepackage{algorithmic}
\usepackage{epstopdf}
\usepackage{subfig}
\usepackage[dvipsnames]{xcolor}

\makeatletter
\let\NAT@parse\undefined
\makeatother
\usepackage{hyperref}
\usepackage{float}
\graphicspath{{figures/}}
\title{\LARGE \bf Manipulation Motion Taxonomy and Coding for Robots}

\author{David Paulius, Yongqiang Huang, Jason Meloncon, and Yu Sun
\thanks{
The authors are members of the Robot Perception and Action Lab (RPAL) in the Department of Computer Science \& Engineering at the University of South Florida, Tampa, FL, USA. Jason Meloncon is a former undergraduate researcher.
(Contact email: \texttt{yusun@mail.usf.edu)}}
}

\begin{document}

\maketitle

\thispagestyle{empty}
\pagestyle{empty}

\begin{abstract}
	% RAM paper on manipulations in cooking
	% 1. how many motions are there (David), introduction, foon, videos, 2.5 pg
	% 2. frequent motions (David), 0.5 pg
	% 3. taxonomy of them (David), 1 pg
	% 4. motion types using motion classifier (Jason), 1 pg
	% 5. physical interactive using force (Jason) 1 pg
	% 6. objects are used, functionalities (David) 0.5 pg
	% 7. what are important (David)
	% 8. conclusion 1
%In order to perform human-like manipulations well, we need to understand the intricacies of such motions and activities. 
%In household activities, especially cooking, it is important that we perform manipulations masterfully, and we ensure that motion learning is as efficient as possible.
This paper introduces a taxonomy of manipulations as seen especially in cooking for 1) grouping manipulations from the robotics point of view, 2) consolidating aliases and removing ambiguity for motion types, and 3) provide a path to transferring learned manipulations to new unlearned manipulations.  Using instructional videos as a reference, we selected a list of common manipulation motions seen in cooking activities grouped into similar motions based on several trajectory and contact attributes. Manipulation codes are then developed based on the taxonomy attributes to represent the manipulation motions. The manipulation taxonomy is then used for comparing motion data in the Daily Interactive Manipulation (DIM) data set to reveal their motion similarities. 
% Finally, we discuss how we can deduce affordances from our knowledge representation to identify objects with similar functionality.
% }
\end{abstract}

\section{Introduction}
%Human activities, no matter how complex or trivial, are comprised of manipulations (usually with objects or tools) in a variety of ways to accomplish a task oachieve a certain gr oal.
%That is to say: if we want to understand how to perform human activities in robotics, we must focus on how we can replicate human motions and manipulate objects used in common activities of daily living (ADL).
We aim to build robots which can not only work among human beings safely but robots that can perform tasks as well as humans.  In the development of household assistant robots and other similar technologies, the fluidity of motions is important for the users and developers alike. Learning human-like manipulations has been the objective of reinforcement learning and robot learning for motion generation.
To represent motion from a robotics point of view, in this paper, we introduce a manipulation taxonomy that considers the robotics (i.e.  mechanics) of human manipulations, particularly in the attributes of contact type and trajectory type in cooking activities. Those attributes are directly associated with trajectory generation and control.  It is a representation that a robot could ``understand'' and execute. 

Grasp taxonomies have been extremely inspirational and useful in robotic grasp planning and analysis. A number of works have defined different grasp taxonomies or grasp types \cite{cutkosky1989grasp, worgotter2013simple,bullock2013hand,feix2016grasp,nakamuracomplexities,dai2013functional,abbasi2016grasp,marino2016data} from either videos or grasping data.  Those studies have focused on uncovering more than the dichotomy between power and precision grasps, and they go deep into the way fingers secure objects contained within the hand. Grasping taxonomies have greatly aided in grasp planning for robot manipulation \cite{lin2014grasp,lin2015robot,cini2019choice}. To some degree, this relates to the theory of affordance \cite{Gibson_1977} where we can infer the functionality of an object based on properties of the object itself. Using grasps, we can identify the type of activity happening in a scene, even if the tool is occluded from view, because the type of the grasp can suggest the type of tool being held or manipulated \cite{helbig2010action}. 

However, there is a lack of a manipulation motion taxonomy that focuses on the mechanics of motions -- trajectory and contact in the manipulations. Different from the grasp taxonomy that focuses on the finger kinematics, we prioritize contact and motion trajectory. A mechanics-based manipulation motion taxonomy could help roboticists to consolidate motion aliases, words or expressions of the same or similar motions in terms of mechanics and to eventually use this knowledge for motion generation, analysis, and recognition. A good manipulation motion taxonomy also plays an important role in transferring or generalizing skills learned for one manipulation to others using common attributes.

Using the attributes defining the manipulation taxonomy, we can code manipulation motions using binary-encoded strings, which can represent manipulation in a way that robots can ``understand'' and use to plan and execute. With such strings, we can also consolidate aliases or terms for different motions (even in other languages) since they will be represented in a format that describes the motions on a functional level. A properly represented motion in a machine language is crucial for manipulation knowledge representation \cite{paulius2019survey} such as functional object-oriented network (FOON) \cite{foonet}. Using our proposed motion taxonomy, motions with different names such as ``insert'' and ``pierce'' are represented with the same manipulation code, as they share the same motion and tactile features in the taxonomy. 
%, different motion types will be translated into a machine language that defines them from the point-of-view of robotic manipulation.
%This can be likened to a deep neural network, which takes motions as input and produces a specific output in the form of a binary-encoded string.
%In addition, a taxonomy is not only limited to activity recognition from demonstrations but in developing a roadmap for learning different manipulations since they will typically be learned one by one.
% }

In designing the manipulation motion taxonomy, we first identify motion and contact features that can be used for distinguishing motions. These features are selected based on common characteristics used in robot motion generation and control such that the motions with the same taxonomy features would have the same motion generation and control strategy. In identifying motion codes in the taxonomy, we have taken two different approaches: 1) clustering motions based on intuition according to the motion features, and 2) clustering motions based on real data and experimental observations.

\begin{figure}[t]
	\centering
	\includegraphics[trim=0.25cm 1.5cm 1.5cm 1.5cm,clip, width=\columnwidth]{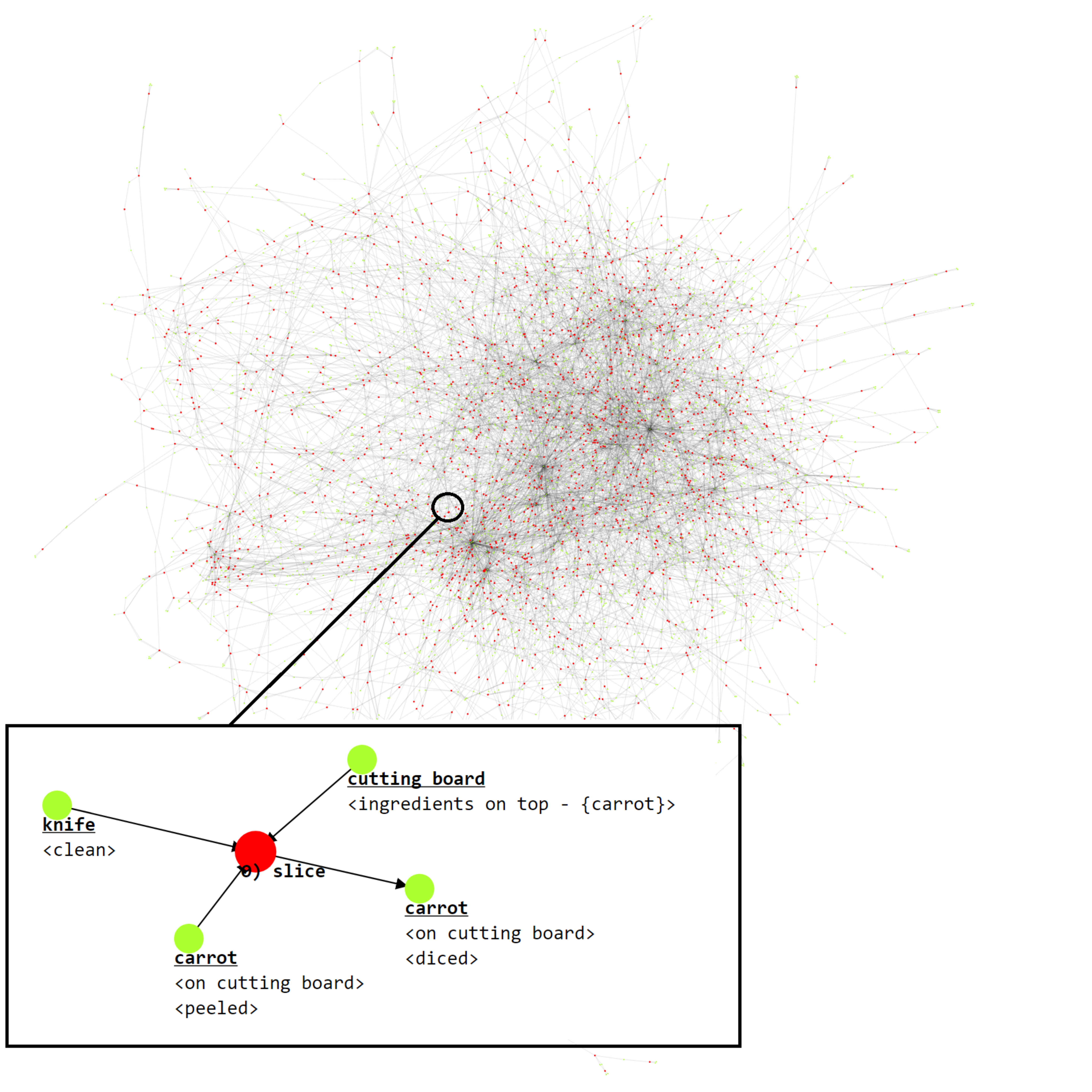}
	\caption{
% 	{\color{blue}[C-1.3] / [C-4.3] 
    An illustration of the universal FOON \cite{foonet}.}
    % }
	\label{fig:universal}
\end{figure}

\section{Manipulation Data}
\label{sec:foon}
In this paper, we focus on manipulation motions as typically observed in cooking activities. The primary sources of video data used in this paper are two sets of instructional videos and their labels. The first set of videos are from the openly available {\it functional object-oriented network} \cite{Paulius2016,Paulius2018}. 
This knowledge representation, inspired by our previous work \cite{Ren2013,lin2015robot}, combines object and motion annotations from 100 instructional videos of cooking activities. We use motion labels from FOON as well as those from our Daily Interactive Manipulation (DIM) data set for our taxonomy.
% It has object and motion labels of 65 YouTube cooking videos. The other set of videos are from the MPII Cooking Activities data set \cite{MaxPlankIICooking} which consists of 273 household activity videos and their annotation (viz. object names and motion types at different timestamps). From these sources, we identify common labels of manipulation motions from cooking activities. 

The object and motion annotations are represented in a directed acyclic graph called a functional object-oriented network (FOON).  The graph contains a combination of object nodes and motion nodes in structures referred to as functional units, which describes a series of procedures needed to create different meals.  We have used FOON for task planning as well as video understanding \cite{jelodar2018long}.
Presently, our FOON combines a total of 100 video demonstrations: 75 videos from YouTube (10 additional videos in addition to the 65 videos from \cite{Paulius2018}), 18 from Activity-Net \cite{caba2015activitynet}, and the remaining 7 from EPIC Kitchens \cite{Damen2018EPICKITCHENS}.
This network (shown as Figure \ref{fig:universal}) contains a total of 5332 nodes, which is made up of 3448 object nodes and 1884 motion node instances.

\begin{figure}[t]
	\centering
    	\includegraphics[trim=1cm 3.25cm 1.25cm 3.25cm,clip,width=\columnwidth]{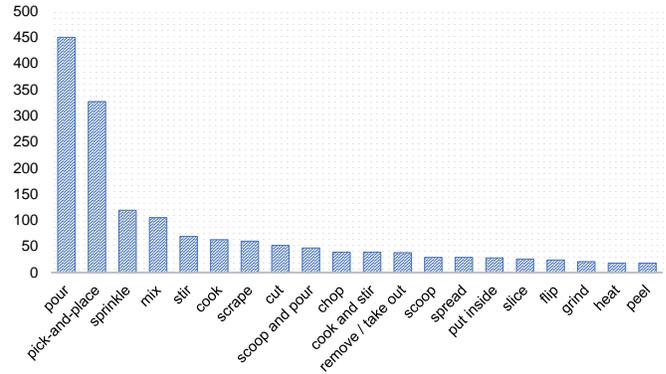}
	\caption{Graph showing the 20 most frequently appearing motions in our universal FOON.
    %   {\color{violet}[C-5.2] 
    % This graph gives the percentage of functional units with a given motion label.
    }
    % }
	\label{fig:motion-freq-merged}
\end{figure}

An indicator of important motions for human activities can be obtained from counting the frequency of motions that appear in the network.
These important motions indicate manipulations which would especially need to be mastered by a robotic system for performing cooking tasks since they are used quite often in cooking.
By identifying such motions, we can give special attention to learning said motions and fine-tuning their performance.
To measure the frequency, we simply count the number of motion node instances (where there is one node per functional unit) in the entire universal FOON since there can be (and there are) multiple instances of each motion node type.
We show the frequency (as percentages) of the top 20 motion types (making up 85\% of all motion nodes) in the universal FOON as Figure \ref{fig:motion-freq-merged}.

One major challenge we have encountered during the process of annotating FOONs is inconsistency among labels used by annotators and among different data sets. In the case of our universal FOON, for example, with multiple annotators, there is always a concern of labels provided by volunteers for describing activities in demonstration videos. To fix this, we would need to review the labels and correct them to match the appropriate motion labels. We also encountered this problem of inconsistency when merging information from other data sets such as the MPII Cooking Activities Dataset \cite{MaxPlankIICooking}, which use different expressions to describe their activities to ours. This difficulty partially motivated us to develop a motion representation that is meaningful to robots.

In many manipulations, contact is a very important component. However, videos can only provide the visual information of the manipulations. It is impossible to analyze the contact characteristics between the objects in manipulations solely using visual features. Therefore, we performed common physical interactive manipulation motions in our lab and collected the interactive force and torque readings along with the motions during the manipulations.  The data set of 32 manipulation types of 3,000 manipulation trials is openly available through \cite{dim}. 

\begin{table*}[h!]
\centering
\caption{
    An outline of the attributes used for representing a motion's mechanics when considering both trajectory and contact features. The attributes form a taxonomy that can group manipulation motions based on their trajectory and contact. Each attribute is also assigned one or two digits of binary code. 
	The codes are then combined in order to create a machine representation of the motion; for example, motion code {\it \{10111010\}} means that we have a contact motion (1) of a rigid engagement type (0) which causes a movement of the acted-upon passive object (11), it has a prismatic but non-revolute trajectory motion (1 and 0), a continuous contact type (1), and the motion is typically unimanual (0). 
}
\label{table:crit}
\begin{tabular}{|p{3.3cm}|p{13cm}|}
\hline
\textit{\textbf{Manipulation Attributions}} 	& \textit{\textbf{Description of Attributes}}	\\ \hline
Contact Type    &	\vspace{-1.5mm}
                        \begin{itemize}   
								\item{0 : \textbf{Non-contact} -- there is little to no contact between active and passive items.}
								\item{1 : \textbf{Contact} -- there is contact between active and passive items.}                          	\end{itemize} \\ \hline
Engagement Type		& \vspace{-1.5mm}	
                    \begin{itemize} 
                        	\item{0 : \textbf{Rigid engagement} -- tool (active) and object(s) (passive) do not change in state or structure.}
                        	\newline 
                        	Rigid engagement sub-classes:
                          	\begin{itemize}
                                \item{00 : \textbf{Stationary} -- there is no movement of the passive item from the action.} 	
                                                   	\item{11 : \textbf{Moving} -- passive object is moved as a result of the manipulation.}   	\end{itemize}

                         	\item{1 : \textbf{Soft engagement} -- the manipulation causes change in the state of tools (active) or objects (passive).}
                         	\newline
                         	Soft engagement sub-classes:
                            \begin{itemize}
%                                      \item{0. \textbf{Self-deforming} - the manipulation involves active deforming when in contact with other objects; the grasped tool changes in some way.}
                                     \item{00 :  \textbf{Admitting/Penetrative} -- contact or action allows for penetration, or the manipulated object is permeable in some way for the tool to enter.}
                                     \item{1 : \textbf{Deforming} -- the manipulation causes deforming, either to the:}
                                     \begin{itemize}
                                      \item{0 : \textbf{Manipulator (active deforming)} -- deformation of active tool}
                                      \item{1 : \textbf{Manipulatee (passive deforming)} -- deformation of passive object(s)}
	                           	\end{itemize}
                          	\end{itemize}
                            \end{itemize} 													\\ \hline
Trajectory Type		&    \vspace{-1.5mm}		\begin{itemize}
                                	\item {(0 -- False, 1 -- True) \textbf{Prismatic} -- the movement trajectory is on a line, plane or surface.}
                                    \item {(0 -- False, 1 -- True) \vphantom{1} \textbf{Revolute} -- the movement is about axes of rotation; the trajectory moves in its orientation domains.}
                              	\end{itemize} \\ \hline
Contact Duration \newline (between tool and objects)			& \vspace{-1.5mm}	
							\begin{itemize} 
                          		\item{0 :  \textbf{Discontinuous} --  active tool or object makes inconsistent contact with the passive object(s).}
                          		\item{1 :  \textbf{Continuous} -- active tool or object makes constant contact with the passive object(s).}      \end{itemize} 	\\ \hline
%                             	\item{1. \textbf{Discontinuous} - tool or object is used in a discontinuous manner where force is exerted instantaneously}
%                                 \begin{itemize}
%                                   	\item {0. \textit{Linear impact} - forceful impact linearly on the object by a tool or the force is applied to the tool.}
%                                    	\item {1. \textit{Angular momentum} - forceful impact causes some rotational moment on the tool or object.}
%                                 \end{itemize}
% 							\end{itemize} 													\\ \hline
Manual Operation     &	\vspace{-1.5mm}
                        \begin{itemize} 
                          		\item{0 : \textbf{Unimanual} -- the action uses a single hand mainly.}
                            	\item{1 : \textbf{Bimanual} -- the action uses two hands to manipulate the tool.}
							\end{itemize} 													\\ \hline
\end{tabular}
\end{table*}

\section{Motion Taxonomy}
\label{sec:tax}
To capture the mechanics of the manipulation motion, we look at the motion from the following main aspects: contact type, engagement type, and trajectory type. We then add two additional aspects that could be useful for planning: contact duration and manual operation (whether unimanual or bimanual) for finer manipulation details. We combine them into a manipulation code to represent a motion. In Table \ref{table:crit}, we describe these attributes in detail. 

\subsection{Motion Attributes}
We mainly distinguish manipulations as {\it contact} or {\it non-contact} motions. Contact motions are those in which there is an interaction between objects, tools or utensils in the demonstration, while non-contact are those in which there is little to no contact.  {\it Contact} motions are those manipulations that involve forces being applied on an object (or a set of objects) where the force is exerted by a tool, utensil or another object. We refer to the tool or utensil as the {\it active} participant in the motion, while objects being acted upon are referred to as {\it passive} participants.  For instance, a hammer exerts force as repeated single, powerful impacts on a nail for the hammering motion, while a softer force can be observed with motions like mixing liquids in a container or brushing a surface with a brush.  In some cases, the robot's hand acts as the active tool in manipulations such as picking-and-placing, squeezing or folding.
We can also have a {\it non-contact} motion type, which will involve the manipulation of tools that make little to no contact on participating passive objects. For instance, when we pour a liquid into a bowl from a cup, the cup does not touch the bowl in a typical pouring action. It is important to note that in pouring, we do not consider the hand gripping the object as a tool.
% Hence, we group pick-and-place as a non-contact motion.

%To classify manipulation based on contact, we have established a set of criteria based on the movement pattern in the action and the force contact a tool makes with other objects.

\begin{figure*}[t]
	\centering
     \includegraphics[trim= .75cm .75cm .75cm 0.75cm, clip, width=\textwidth]{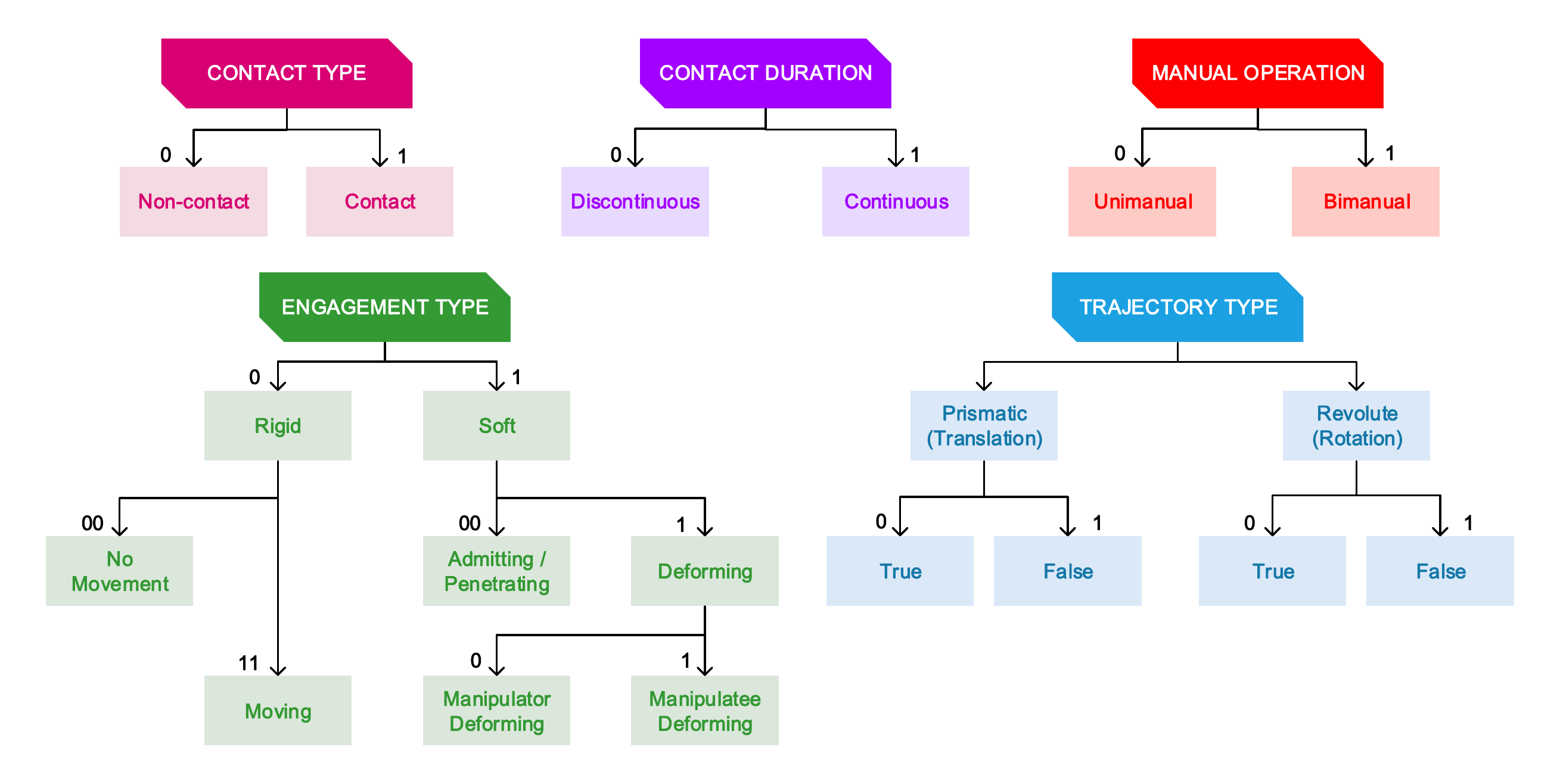}
    \caption{An illustration of different motion attributes as described in Table \ref{table:crit} as hierarchical trees. }
    \label{fig:tax_image}
\end{figure*}

A manipulation motion can also be identified by how an active object {\it engages} with other passive objects.
We identify motion engagement types as either being {\it soft} or {\it rigid}.  {\it Soft engagement} motions are those where either the active tool or the passive object undergoes a change in its shape from contact with each other.  {\it Rigid} or {\it neutral engagement} motions have neither the tool nor the objects change in their shape, state or form as a result of direct contact. However, these motions can either cause some sort of {\it movement} in the manipulation or the object being acted upon does {\it not move} from the manipulation. For instance, with a spatula, one can pick up items without changing the physical state of the manipulator tool and the manipulated object, but the passive item would be moved from one location to another.

Soft engagement contact can be broken down into three subcategories: 1) {\it admitting} or {\it penetrative}, where the tool can penetrate the object without deformation of the tool and the passive object allows the tool to enter it, or 2) {\it deforming}, where either the active or passive object deforms in some way.  The latter can be further broken down into either deforming of the {\it manipulator}, where the active tool itself changes in its shape or deforms for manipulation upon an object, or deforming of the {\it manipulatee}, where the passive object changes in its state or shape and the active tool remains rigid and does not deform.  As an example of an admitting engagement, when scooping flour from a bowl, the spoon or cup penetrates the ingredients.  A manipulator-deforming engagement type can be observed when using a brush, for instance, since the bristles will bend and deform in shape from the default appearance of a brush.  As for a manipulatee-deforming engagement such as cutting, the active knife deforms the passive object by changing its shape from its natural state to pieces for the purpose of cooking.

With contact made between the active tool and the passive object, engagement can either be {\it continuous}, where there is a constant interaction or force in the manipulation over the duration of the action, or {\it discontinuous}, where there is little to no constant or non-persistent contact between them.
Discontinuous motions tend to be those which can be identified by sharp periods of force.
For example, in the case of pick-and-place, the only contact between the object and the environment in the pick-and-place process are at the beginning and the end of the process -- breaking and establishing contact between the picked object and the support environment. However, since the hand is considered to be the active tool, which continuously grips the object, this action is considered as continuous contact.
With an action such as dipping, the object will only make temporary contact with contents usually held within a container.

As for manipulation motion types, the movement can be {\it prismatic}, where it undergoes linear translation across a line/plane (e.g. cutting is usually a vertical motion in 1D), or it can be {\it revolute} or {\it rotational}, where the object or tool undergoes a change in orientation and it moves about axes of rotation (e.g. pouring typically involves the rotation of a cup to allow liquid to flow into a receiving container).
Manipulation motions are not confined to a single trajectory type since certain manipulations combine rotation and translation; hence, these two subcategories are not mutually exclusive. An example of this type of motion is folding.

Finally, these motion types can also be described by the number of hands (or end-effectors) regularly used in the action.
We can classify them as {\it unimanual} (involving one hand) or {\it bimanual} (involving both hands) in terms of manipulation of the active tool or item.
Sprinkling salt from a shaker can be considered as a unimanual action since we can hold the shaker and shake it with one hand, while rolling or flattening is usually a bimanual action since a rolling pin requires two hands to operate.
This criterion is important for determining which motions we can execute since some robotic systems are not built consistently to human anatomy (i.e. with two arms, two hands, and similar joints).

Figure \ref{fig:tax_image} illustrates the manipulation taxonomy described in Table \ref{table:crit} as five hierarchical trees. Each manipulation motion will be grouped according to the taxonomy trees and assigned a string of binary manipulation code. 
The binary string is a combination of manipulation attributes in the following order from left to right: contact type, engagement type, trajectory type, contact duration and manual operation.

\subsection{Manipulation Codes}
Based on the taxonomy, each motion type can be represented with a {\it manipulation code} which can be used for representing each motion as detailed in our taxonomy. 
In Table \ref{table:force}, we assigned manipulation codes to common cooking motions as seen in both FOON and DIM.
Several motions end up naturally clustered because of common codes.

Mixing/stirring is assigned the same code as inserting/piercing since they are both admitting actions, have prismatic trajectories, and they are classified as continuous contact motions.
Cutting/slicing/chopping along with motions such as mashing, rolling (unimanual), peeling, shaving, and spreading are clustered together mainly because of their manipulatee-deforming and prismatic properties. 
This group is separate to that containing pulling apart and grating because they are typically bimanual actions.

\begin{table}[t]
\centering
\caption{Manipulation Code (based on criteria in Table \ref{table:crit}). 
        Refer to the index in Table \ref{table:crit} or Figure \ref{fig:tax_image} for the meaning behind binary digits.
		}
\label{table:force}
\begin{tabular}{|l|p{5cm}|}
\hline
\textit{\textbf{Manipulation Code}} & \textit{\textbf{Motion Types}}  \\ \hline
% 00001001	&	gather      \\ \hline
00000100	&   shake/sprinkle	\\ \hline
00001000  	&   rotate, pour    	\\ \hline
10111000	&	poke        \\ \hline	
10111010    &   pick-and-place, push (rigid)	\\ \hline
10111100	&	flip        \\ \hline		
11001000	&	dip			\\ \hline
11001010	&	insert, pierce, mix, stir		\\ \hline
11001100	&	scoop 		\\ \hline	
11101010	&	brush, wipe, push (deforming)   \\ \hline	
11110100	&	tap, crack (egg)    \\ \hline
11110111	&	twist (open/close container) \\ \hline 
11111010	&	cut, slice, chop, mash, roll (unimanual), peel, scrape, shave, spread, squeeze, press, flatten 	\\ \hline
11111011	&	roll (bimanual), pull apart, grate 			\\ \hline
11111110	&	fold (wrap/unwrap) \\ \hline
\end{tabular}
\end{table}

% \begin{figure}[t]
% 	\centering
% 	\includegraphics[width=7cm]{fig_taxonomy-3.png}
% 	\caption{ 
% % 	{\color{violet}[C-5.3] 
% 	    A basic way of classifying motions.}
%     % }
% 	\label{fig:taxx}
% \end{figure}

\section{Classifying Motions with Real Data}
\label{sec:kl}
We have established a motion taxonomy for grouping motions which are similar to one another based on force and motion using attributes such as contact versus non-contact.
In this section, we support our taxonomy by comparing force reading data for different motion types.
As we described in Section \ref{sec:foon}, several demonstrations were recorded using position/orientation and force sensors for a variety of human activities and are featured in the DIM data set.
The objective here is to match each activity to a motion type and to determine whether the measurements show that certain motion types are alike to other motion types, thus determining whether the clusters from Table \ref{table:force} aligns with real data.

\subsection{Finding Motion Similarity}
The DIM dataset is the only data set at the moment that contains contact 6-axis force data of many manipulation motions \cite{huang2016recent, huang2018dataset}. However, due to the limitation of the force sensor in its data collection setup, it does not have manipulations involving high force or torque, such as squeezing, mashing, or pressing. 
Additionally, we did not analyze non-contact motions (such as pouring or sprinkling/shaking) because there are no interactive forces to measure between active and passive objects.
It is for that reason we do not have mappings to all motion clusters.
Several motions were collected as multiple variations of demonstrations, and so we try to combine all recordings in this data set.

\begin{figure}[t]
	\centering
	\includegraphics[trim=0.5cm 1.75cm 2.5cm 2.25cm,clip, width=\columnwidth]{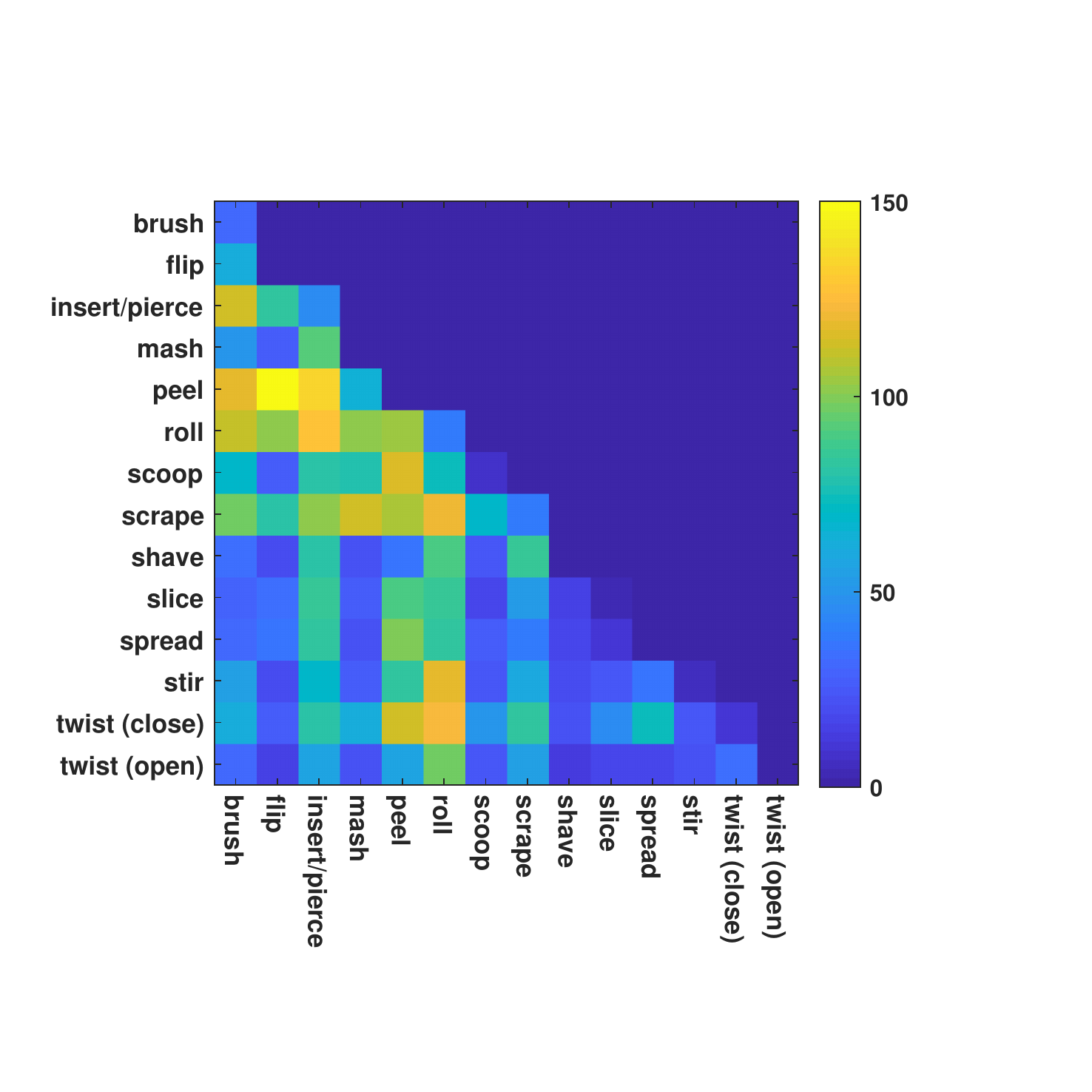}
	\caption{Matrix showing the Kullback-Leibler divergence values as computed from our manipulation motions data set.
	We only show the lower diagonal since the computed matrix is symmetric (this image is best viewed in colour). 
% 		We compress the matrix by averaging the KL values for various motion type instances
% 		; we show the non-averaged matrix as Figure \ref{fig:kl_div2}
		.
}
	\label{fig:kl_div1}
\end{figure}

Using the force data, we created a representative model for each motion type using Gaussian Mixture Models (GMM).  Each GMM represents a force distribution across space to derive a motion description of a motion type, and they are built by combining the data points generated in multiple trials of demonstrations. To measure the similarity of motions using their individual force distributions, we use the Kullback-Leibler (KL) divergence method \cite{kullback1951information}.
The typical method for measuring KL divergence between two distributions is to use random sampling between different points; however, this is a very intensive task for us to do with GMMs, and so we used the variational approximation of KL divergence (as proposed in \cite{hershey2007approximating}) as the distance measure between a pair of different motions.
Originally, this metric is asymmetric and it is non-transitive (i.e. the KL divergence value from A to B will not be the same as that from B to A).
However, we can obtain a symmetric result by taking the average of the divergence values obtained from the two sets of pairs (i.e. we take the value from A to B and B to A and computing the average).
Since we have multiple recordings for certain motion types, we also computed the average of all KL divergence values computed for each of those instances.
This makes it easier to interpret the pairwise values we obtain, which we present in a matrix form as Figure \ref{fig:kl_div1}.
% The deeper the blue, the more similar two motions are to one another.
The values obtained from KL divergence are unbounded and non-negative, where the closer the value is to 0 (based on color, the deeper the shade of the blue), the more two distributions are considered to be alike; conversely, the larger the value obtained from this calculation (based on colour, the lighter the shade of yellow), the more dissimilar two manipulation motion types are from one another based on force readings. Matrix values are symmetric, so we omitted the upper diagonal values.

The main question we will be addressing in this section is: how well do our motion clusters match real supporting data?
We determine this by looking at how similar motions classified as certain clusters match up to others that are also considered to be in the same cluster based on force/torque readings.
In Figure \ref{fig:kl_div1}, we have certain activity pairs whose motion labels agree with our taxonomy such as: 
mashing to slicing, mashing to shaving, spreading to shaving, spreading to mashing, peeling to shaving, and twisting for both directions. There are several motions which are close to one another but differ to the clusters in Table \ref{table:force} due to one or two attributes.
Even though brushing and shaving are considered different in the taxonomy, this is only due to the nature of the tools; brushing is considered to be manipulator-deforming, while shaving is manipulatee-deforming.
The movement type and force application are expected to be similar aside from the deformation type found in these tools, and therefore these motions can be considered to be similar.
%Everything else about these motions are similar according to their motion coding.
Similarly, flipping and scooping are similar to one another because they are both prismatic and revolute; however, flipping is considered as a rigid engagement motion, while scooping is an admitting, soft engagement motion.
Inserting/piercing is considered to be somewhat distant to all other motions, with perhaps the closest to twisting, which does not match our expectations.

Other pairs which we expected to be similar but they did not have low KL divergence values include peeling and scraping; conversely, motion pairs that were deemed similar but do not match our taxonomy include flipping and mashing, flipping and shaving, stirring to slicing, stirring to shaving, and stirring to spreading.
Twisting open is found to be similar to many other motions such as slicing and shaving which are not revolute but prismatic only motions.
This illustrates that these features should not be neglected when comparing motion data.
Since the KL divergence only considers force readings, we neglect other factors which may give away unlikely matching candidates, which are likely to be obtained from an analysis of motion trajectory data or video analysis. This is why some similarities do not match with the intra-clustering of motions.

\section{Conclusion}
\label{sec:con}
In conclusion, our aim in this paper was to investigate the robotic attributes of manipulation tasks as seen in cooking and use them to create an effective representation of manipulation motions.  By identifying a motion taxonomy, we were able to assign binary-encoded strings, which we called manipulation codes, that describe attributes of a particular motion based on trajectory and contact properties.
Manipulation codes can be used to determine motion types that are similar to one another. The taxonomy and codes allow researchers to represent and group manipulations from the robotics point of view. In addition, by representing motions as manipulation codes, we can effectively consolidate aliases (or different labels or words) thus removing ambiguity among motion types. Moreover, comparing the codes between manipulations provides a path towards transferring learned manipulations to new unlearned manipulations.

To show that the motion code assignments given to different motion types hold up in measuring similarity (or dissimilarity) between motion types, we performed experiments using collected demonstration data.
We showed that the force reading data for certain motion types naturally cluster with other motion types, supporting the taxonomical clusters described in the paper.  For a better measure of similarity and support for the taxonomy, we would need to collect force data for other motions that we did not include in the analysis. Furthermore, we may also identify other obtainable attributes to be included within the taxonomy, which can be selected based on the proposed task and available resources.

\section*{Acknowledgement}
This material is based upon work supported by the National Science Foundation under Grants No. 1421418 and 1560761.

\bibliographystyle{IEEEtran}
\bibliography{ref}

\begin{thebibliography}{10}
\providecommand{\url}[1]{#1}
\csname url@samestyle\endcsname
\providecommand{\newblock}{\relax}
\providecommand{\bibinfo}[2]{#2}
\providecommand{\BIBentrySTDinterwordspacing}{\spaceskip=0pt\relax}
\providecommand{\BIBentryALTinterwordstretchfactor}{4}
\providecommand{\BIBentryALTinterwordspacing}{\spaceskip=\fontdimen2\font plus
\BIBentryALTinterwordstretchfactor\fontdimen3\font minus
  \fontdimen4\font\relax}
\providecommand{\BIBforeignlanguage}[2]{{%
\expandafter\ifx\csname l@#1\endcsname\relax
\typeout{** WARNING: IEEEtran.bst: No hyphenation pattern has been}%
\typeout{** loaded for the language `#1'. Using the pattern for}%
\typeout{** the default language instead.}%
\else
\language=\csname l@#1\endcsname
\fi
#2}}
\providecommand{\BIBdecl}{\relax}
\BIBdecl

\bibitem{cutkosky1989grasp}
M.~R. Cutkosky, ``On grasp choice, grasp models, and the design of hands for
  manufacturing tasks,'' \emph{IEEE Transactions on robotics and automation},
  vol.~5, no.~3, pp. 269--279, 1989.

\bibitem{worgotter2013simple}
F.~W{\"o}rg{\"o}tter, E.~E. Aksoy, N.~Kr{\"u}ger, J.~Piater, A.~Ude, and
  M.~Tamosiunaite, ``A simple ontology of manipulation actions based on
  hand-object relations,'' \emph{IEEE Transactions on Autonomous Mental
  Development}, vol.~5, no.~2, pp. 117--134, 2013.

\bibitem{bullock2013hand}
I.~M. Bullock, R.~R. Ma, and A.~M. Dollar, ``A hand-centric classification of
  human and robot dexterous manipulation,'' \emph{IEEE transactions on
  Haptics}, vol.~6, no.~2, pp. 129--144, 2013.

\bibitem{feix2016grasp}
T.~Feix, J.~Romero, H.-B. Schmiedmayer, A.~M. Dollar, and D.~Kragic, ``{The
  GRASP taxonomy of human grasp types},'' \emph{IEEE Transactions on
  Human-Machine Systems}, vol.~46, no.~1, pp. 66--77, 2016.

\bibitem{nakamuracomplexities}
Y.~C. Nakamura, D.~M. Troniak, A.~Rodriguez, M.~T. Mason, and N.~S. Pollard,
  ``The complexities of grasping in the wild,'' in \emph{2017 IEEE-RAS 17th
  International Conference on Humanoid Robotics (Humanoids)}.\hskip 1em plus
  0.5em minus 0.4em\relax IEEE, 2017, pp. 233--240.

\bibitem{dai2013functional}
W.~Dai, Y.~Sun, and X.~Qian, ``Functional analysis of grasping motion,'' in
  \emph{Intelligent Robots and Systems (IROS), 2013 IEEE/RSJ International
  Conference on}.\hskip 1em plus 0.5em minus 0.4em\relax IEEE, 2013, pp.
  3507--3513.

\bibitem{abbasi2016grasp}
B.~Abbasi, E.~Noohi, S.~Parastegari, and M.~{\v{Z}}efran, ``Grasp taxonomy
  based on force distribution,'' in \emph{Robot and Human Interactive
  Communication (RO-MAN), 2016 25th IEEE International Symposium on}.\hskip 1em
  plus 0.5em minus 0.4em\relax IEEE, 2016, pp. 1098--1103.

\bibitem{marino2016data}
H.~Marino, M.~Gabiccini, A.~Leonardis, and A.~Bicchi, ``Data-driven human grasp
  movement analysis,'' in \emph{ISR 2016: 47st International Symposium on
  Robotics; Proceedings of}.\hskip 1em plus 0.5em minus 0.4em\relax VDE, 2016,
  pp. 1--8.

\bibitem{lin2014grasp}
Y.~Lin and Y.~Sun, ``Grasp planning based on strategy extracted from
  demonstration,'' in \emph{2014 IEEE/RSJ International Conference on
  Intelligent Robots and Systems}.\hskip 1em plus 0.5em minus 0.4em\relax IEEE,
  2014, pp. 4458--4463.

\bibitem{lin2015robot}
------, ``Robot grasp planning based on demonstrated grasp strategies,''
  \emph{The International Journal of Robotics Research}, vol.~34, no.~1, pp.
  26--42, 2015.

\bibitem{cini2019choice}
F.~Cini, V.~Ortenzi, P.~Corke, and M.~Controzzi, ``On the choice of grasp type
  and location when handing over an object,'' \emph{Science Robotics}, vol.~4,
  no.~27, p. eaau9757, 2019.

\bibitem{Gibson_1977}
J.~Gibson, ``The theory of affordances,'' in \emph{Perceiving, Acting and
  Knowing}, R.~Shaw and J.~Bransford, Eds.\hskip 1em plus 0.5em minus
  0.4em\relax Hillsdale, NJ: Erlbaum, 1977.

\bibitem{helbig2010action}
H.~B. Helbig, J.~Steinwender, M.~Graf, and M.~Kiefer, ``Action observation can
  prime visual object recognition,'' \emph{Experimental Brain Research}, vol.
  200, no. 3-4, pp. 251--258, 2010.

\bibitem{paulius2019survey}
D.~Paulius and Y.~Sun, ``A survey of knowledge representation in service
  robotics,'' \emph{Robotics and Autonomous Systems}, vol. 118, pp. 13--30,
  2019.

\bibitem{foonet}
``{FOON Website:} {Graph Viewer} and {Videos},'' \url{http://www.foonets.com},
  accessed: July 31, 2019.

\bibitem{Paulius2016}
D.~Paulius, Y.~Huang, R.~Milton, W.~D. Buchanan, J.~Sam, and Y.~Sun,
  ``{Functional Object-Oriented Network for Manipulation Learning},'' in
  \emph{2016 IEEE/RSJ International Conference on Intelligent Robots and
  Systems (IROS)}.\hskip 1em plus 0.5em minus 0.4em\relax IEEE, 2016, pp.
  2655--2662.

\bibitem{Paulius2018}
D.~Paulius, A.~B. Jelodar, and Y.~Sun, ``{Functional Object-Oriented Network:
  Construction \& Expansion},'' in \emph{{ICRA 2018 - IEEE International
  Conference on Robotics and Automation}}, Brisbane, Australia, May 2018.

\bibitem{Ren2013}
S.~Ren and Y.~Sun, ``Human-object-object-interaction affordance,'' in
  \emph{Workshop on Robot Vision}, 2013.

\bibitem{jelodar2018long}
A.~B. Jelodar, D.~Paulius, and Y.~Sun, ``{Long Activity Video Understanding
  using Functional Object-Oriented Network},'' \emph{IEEE Transactions on
  Multimedia}, 2018.

\bibitem{caba2015activitynet}
B.~G. Fabian Caba~Heilbron, Victor~Escorcia and J.~C. Niebles, ``{ActivityNet:
  A Large-Scale Video Benchmark for Human Activity Understanding},'' in
  \emph{Proceedings of the IEEE Conference on Computer Vision and Pattern
  Recognition}, 2015, pp. 961--970.

\bibitem{Damen2018EPICKITCHENS}
D.~Damen, H.~Doughty, G.~M. Farinella, S.~Fidler, A.~Furnari, E.~Kazakos,
  D.~Moltisanti, J.~Munro, T.~Perrett, W.~Price, and M.~Wray, ``{Scaling
  Egocentric Vision: The EPIC-KITCHENS Dataset},'' in \emph{European Conference
  on Computer Vision (ECCV)}, 2018.

\bibitem{MaxPlankIICooking}
\BIBentryALTinterwordspacing
M.~Rohrbach, S.~Amin, M.~Andriluka, and B.~Schiele, ``A database for fine
  grained activity detection of cooking activities.'' in \emph{CVPR}.\hskip 1em
  plus 0.5em minus 0.4em\relax IEEE Computer Society, 2012, pp. 1194--1201.
  [Online]. Available:
  \url{http://dblp.uni-trier.de/db/conf/cvpr/cvpr2012.html#RohrbachAAS12}
\BIBentrySTDinterwordspacing

\bibitem{dim}
``{Daily Interactive Manipulation (DIM) Dataset},''
  \url{http://rpal.cse.usf.edu/datasets_manipulation.html}, accessed: July 31,
  2019.

\bibitem{huang2016recent}
Y.~Huang, M.~Bianchi, M.~Liarokapis, and Y.~Sun, ``Recent data sets on object
  manipulation: A survey,'' \emph{Big data}, vol.~4, no.~4, pp. 197--216, 2016.

\bibitem{huang2018dataset}
Y.~Huang and Y.~Sun, ``A dataset of daily interactive manipulation,'' \emph{The
  International Journal of Robotics Research}, vol.~38, no.~8, p. 879–886,
  2019.

\bibitem{kullback1951information}
S.~Kullback and R.~A. Leibler, ``On information and sufficiency,'' \emph{The
  annals of mathematical statistics}, vol.~22, no.~1, pp. 79--86, 1951.

\bibitem{hershey2007approximating}
J.~R. Hershey and P.~A. Olsen, ``Approximating the kullback leibler divergence
  between gaussian mixture models,'' in \emph{Acoustics, Speech and Signal
  Processing, 2007. ICASSP 2007. IEEE International Conference on},
  vol.~4.\hskip 1em plus 0.5em minus 0.4em\relax IEEE, 2007, pp. IV--317.

\end{thebibliography}

\end{document}